\def\year{2020}\relax
\documentclass[letterpaper]{article} % DO NOT CHANGE THIS
\usepackage{aaai20}  % DO NOT CHANGE THIS
\usepackage{times}  % DO NOT CHANGE THIS
\usepackage{helvet} % DO NOT CHANGE THIS
\usepackage{courier}  % DO NOT CHANGE THIS
\usepackage[hyphens]{url}  % DO NOT CHANGE THIS
\usepackage{graphicx} % DO NOT CHANGE THIS
\usepackage{graphicx}  % DO NOT CHANGE THIS
\usepackage{amssymb}
\usepackage{algorithm}
\usepackage{algorithmic}
\usepackage{amsmath}
\usepackage{subfigure}

\usepackage{multirow}
\newcommand{\tabincell}[2]{\begin{tabular}{@{}#1@{}}#2\end{tabular}}

\nocopyright
\title{GLA-Net: An Attention Network with Guided Loss for Mismatch Removal}
\author{Zhi Chen, Fan Yang, Wenbing Tao
	\thanks{corresponding author}\\
	National Key Laboratory of Science and Technology on Multispectral Information Processing\\
	School of Artifical Intelligence and Automation, Huazhong University of Science and Technology, China\\
	\{hust\_zhichen, hust\_fanyang, wenbingtao\}@hust.edu.cn
}
 
\begin{document}

\maketitle

\begin{abstract}
Mismatch removal is a critical prerequisite in many feature-based tasks. Recent attempts cast the mismatch removal task as a binary classification problem and solve it through deep learning based methods. In these methods, the imbalance between positive and negative classes is important, which affects network performance, i.e., Fn-score. To establish the link between Fn-score and loss, we propose to guide the loss with the Fn-score directly. We theoretically demonstrate the direct link between our Guided Loss and Fn-score during training. Moreover, we discover that outliers often impair global context in mismatch removal networks. To address this issue, we introduce the attention mechanism to mismatch removal task and propose a novel Inlier Attention Block (IA Block). To evaluate the effectiveness of our loss and IA Block, we design an end-to-end network for mismatch removal, called GLA-Net \footnote{Our code will be available in Github later.}. Experiments have shown that our network achieves the state-of-the-art performance on benchmark datasets. 
\end{abstract}

\section{Introduction}
\noindent Establishing stable and abundant feature matches between overlapping image pairs is a fundamental component of many tasks in computer vision, such as Structure from Motion (SfM) \cite{schonberger2016structure}, simultaneous localization and mapping (SLAM) \cite{benhimane2004real} and so on. Due to the ambiguity of local texture information, the matching results often contain a large number of mismatches. Recently, some methods \cite{moo2018learning} adopt deep learning for mismatch removal. Specifically, they first obtain massive putative feature correspondences through hand-crafted local feature descriptors with loose matching conditions, such as SIFT \cite{lowe2004distinctive}. Then the putative set is divided into positives (inliers) and negatives (outliers) through a trainable deep learning network. 
The positive class of network classification is considered as the final matching result.

\begin{figure}[t]
	\centering
	\includegraphics[width=1\columnwidth]{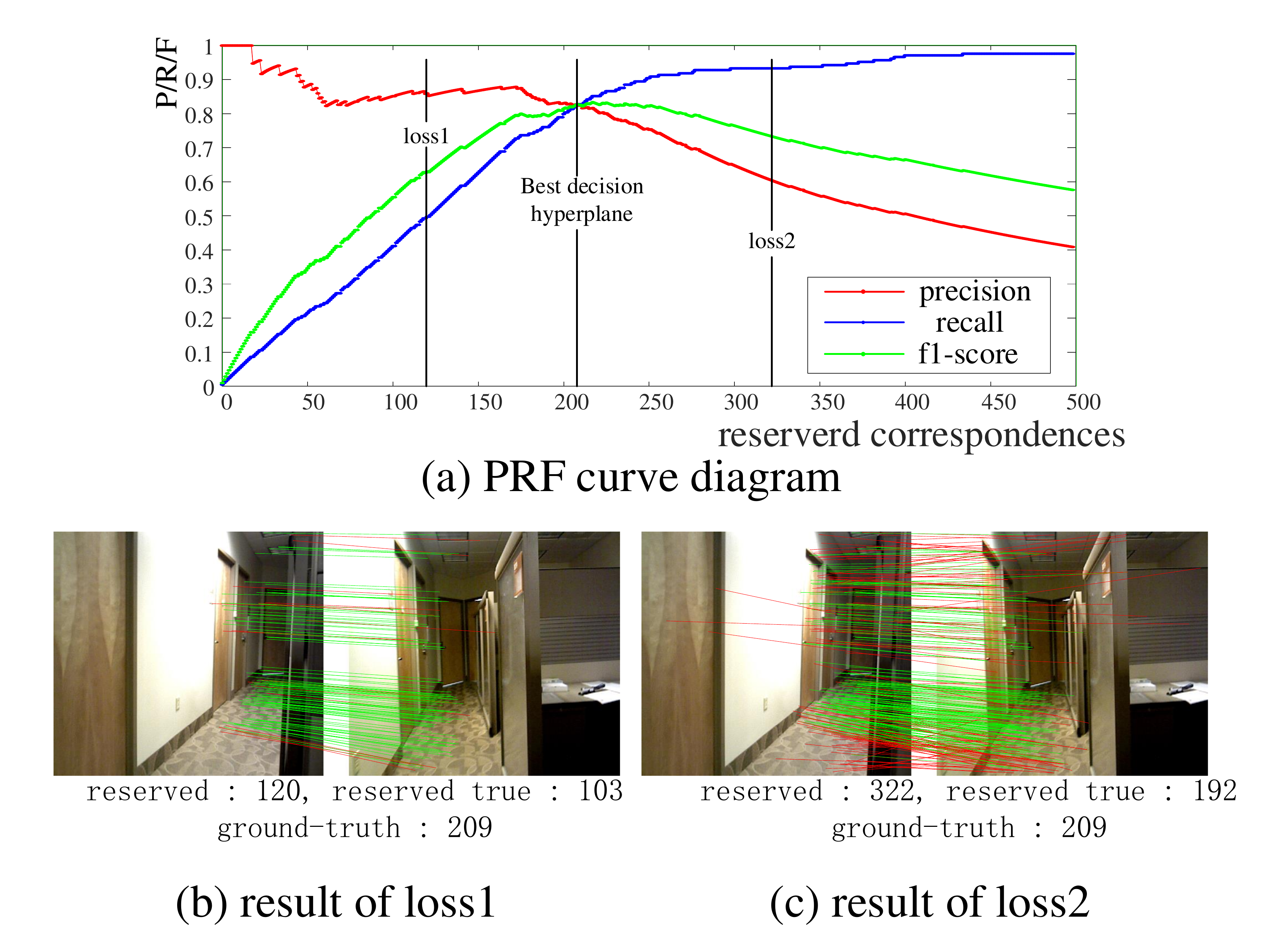} % Reduce the figure size so that it is slightly narrower than the column. Don't use precise values for figure width.This setup will avoid overfull boxes. 
	\caption{The top (a) is the curve diagram of the reserved correspondences and PRF, and the bottom (b, c) is the results of the networks trained by loss1 and loss2 respectively. The red line in (b) and (c) is outliers, while green is inliers. The boundaries of loss1 and loss2 are labeled on (a). \textbf{See text for more details.}} 
	\label{introduction}
\end{figure}

In general, the number of positive and negative instances in mismatch removal task is imbalanced. Indeed, the class imbalance problem has a great impact on the classification results \cite{lin2017focal}. Recent deep learning based mismatch removal networks \cite{moo2018learning,Ploetz:2018:NNN,zhao2019nm} address the class imbalance through the cost sensitive loss. They assign a fixed weight coefficients to the losses of positive and negative classes. The loss weight ratio and the quantitative ratio of the positive and negative classes are mutually reciprocal. However, the fixed weight coefficient usually leads the network to focus too much on a certain class of classification. Fig. \ref{introduction} is a simple example that can show our motivation. We first train a deep learning network with an ordinary cross entropy loss. 
The output of the network is the probability value that each instance belongs to positive class. We sort the output and manually determine how many matches are reserved. Then we can get the curve diagram between the precision (P), recall (R), F1-score (F) and the number of reserved correspondences as shown in Fig. \ref{introduction}.
We define the boundary with the highest F1-score as the \textit{best decision hyperplane}.
After that, we train two networks with two different cost sensitive losses, called loss1 and loss2. The loss1's weights are $1:1$, while loss2's is determined by the quantitative ratio between positive and negative class. As shown in Fig. \ref{introduction}, since the number of negatives on this data set is much larger than positives, the weight of the
negative class in loss1 is too large. Therefore, the network is too biased to the classification accuracy of negative classes, and the classification result of loss1 maintains high precision and low recall. In contrast, the classification result of loss2 has higher recall and lower precision. Both these two loss functions cannot achieve a good trade-off between precision and recall, resulting in a poor Fn-score. To address this issue, we theoretically analyze how to establish a \emph{direct} relationship between loss and Fn-score by automatically adjusting the loss' weights of positives and negatives. Based on these analyses, we propose an algorithm that can guide the loss according to Fn-score.
Specifically, we treat the classification of positive and negative categories as two separate issues. The numerical derivatives of Fn-score with respect to positives and negatives are obtained in each iteration of training, and the weight coefficients are computed through the derivatives. Thus, with the Guided Loss, the network can get closer to the \textit{best decision hyperplane}.

Besides loss function, another important issue is to eliminate the contexts of outliers. In mismatch removal tasks, the wrong correspondences (outliers) have fatal effects on model fitting. Thus, traditional model fitting methods, such as RANSAC \cite{fischler1981random}, often eliminate outliers' effect by obtaining an outlier-free set to fit model through massive sampling. However, recent learning based attempts \cite{zhao2019nm,moo2018learning} ignore the effect of outliers and extract the global context through Context Normalization (CN). It is an undifferentiated operation for each instance, so it cannot filter out any outlier context. To address this issue in deep learning, we introduce the attention mechanism to the mismatch removal networks and propose an Inlier Attention Block (IA Block). It learns an outlier indicating matrix and assigns a corresponding weight to each instance through this matrix. The IA Block reduces the outliers' effects while extracting the global context.

In a nutshell, our contributions is threefold:

%GLA-net is an effective network, which only needs the coordinates of feature correspondences without any additional information.

%Moreover, deep learning networks can introduce outlier context when fitting model. Recent learning based attempts \cite{zhao2019nm,moo2018learning} adopt Context Normalization (CN) for extracting global context. CN treats every instance equally, and normalizes the features by their mean and variance. The undifferentiated global context extraction can bring wrong context into the network. In 
%
%
%Traditional model fitting methods, such as RANSAC \cite{fischler1981random}, often eliminate outlier's impact by obtain a outlier-free set to fit model through random sampling. However, the sampling and selection pipeline is difficult to be performed in deep learning networks. To eliminate the impact of outlier, we introduce the attention mechanism, which are commonly used in video classification tasks, to reduce the weights of outliers. 

%We present a Inlier Attention Block (IA-Block) to fit model in a way of paying attention. 
%
%
%Specifically, IA-Block, a backbone architecture, learns a soft classification result in supervised or unsupervised way for discriminately extracting global context. Finally, the guided loss and IA-Block are integrated into an end-to-end network called GLA-Net in a coarse to fine manner.

\begin{itemize}
	\item A novel Guided Loss is proposed for mismatch removal. It establishes a direct relationship between loss functions and Fn-score. %加一点效果上的描述
	
	\item The attention mechanism is introduced to our network by the proposed IA Block. It somewhat reduces outliers' damage to the global context.%加一点效果，加一些定量的描述
	
	\item An end-to-end network called GLA-Net is presented by means of IA Block and Guided Loss for mismatch removal task. The presented network can achieve the state-of-the-art performance on benchmark datasets with various scenes and proportions of inliers.

	%\item You must use the 2019 AAAI Press \LaTeX{} style file and bib file, which are located in the 2019 AAAI Author Kit.
	%\item You must complete, sign, and return by the deadline the AAAI copyright form (unless directed by AAAI Press to use the AAAI Distribution License instead).
	%\item You must read and format your paper source and PDF according to the formatting instructions for authors.
	%\item You must submit your electronic files and abstract using our electronic submission form \textbf{on time.}
	%\item You must pay any required page or formatting charges to AAAI Press so that they are received by the deadline.
	%\item You must check your paper before submitting it, ensuring that it compiles without error, and complies with the guidelines found in the AAAI Author Kit.
\end{itemize}
\section{Related Works}
\noindent \textbf{Model fitting methods} Model fitting methods usually determine inliers by whether they satisfy the epipolar geometric model. The classic RANSAC \cite{fischler1981random} adopts a hypothesize-and-verify pipeline, so do its variations, such as PROSAC \cite{chum2005matching}, SCRAMSAC \cite{sattler2009scramsac}. Besides, many modifications of RANSAC have been proposed. Some methods \cite{chum2005matching,fragoso2013evsac} propose sampling strategies to reduce sampling frequency or increase sampling stability. Some other methods \cite{chum2003locally,barath2018graph} augment the RANSAC by performing a local optimization step to the so-far-the-best model. However, these methods can not deal with the data with the low ratio of inliers. What's more, some complex models cannot be expressed for a single epipolar geometric model, such as multi-consistency matching \cite{xiao2019superpixel}.

%\noindent \textbf{Non-parametric Methods} 

\noindent \textbf{Learning Based Methods}
Since deep learning has been successfully applied for dealing with unordered data \cite{qi2017pointnet}, learning based methods attract great interest in mismatch removal tasks. %加一个目的
LFGC-Net \cite{moo2018learning} reformulates the mismatch removal task as a binary classification problem. It utilizes a simple Context Normalization (CN) operation to extract global context. Based on CN, some network variations are proposed. NM-Net \cite{zhao2019nm} employs a simple graph architecture with an affine compatibility-specific neighbor mining approach to mine local context. $ \rm N^{3}$-Net \cite{Ploetz:2018:NNN} presents a continuous deterministic relaxtaion of KNN selection and a $\rm N^{3}$ block to mine non-local context. Besides deep learning based methods, LMR \cite{ma2019lmr} constructs local consistency features with a machine learning classifier for mismatch removal. 

%By introducing local constraints, NM-Net achieves better results than LFGC-Net. However, since NM-Net need local affine information to mine neighbor, its application scope is limited and many feature descriptors (such as SIFT) cannot be directly applied. Indeed, local constraints can be replaced by appropriate loss functions. Besides, all of the methods do not pay more attention to the inliers when training the model. 

\noindent \textbf{Class Imbalance} 
The problem of class imbalance has received much attention in object classification \cite{he2016deep} and object detection \cite{ren2015faster}. In object classification, class imbalance are broadly researched by sampling based preprocessing techniques \cite{chawla2002smote}, cost sensitive learning \cite{akbani2004applying}. In object detection, some methods avoid fitting too many simple samples through mining hard examples \cite{shrivastava2016training}. Focal Loss \cite{lin2017focal} down-weights the losses assigned to well-classified examples through reshaping the standard cross entropy loss function. All of the above methods utilize a loss function with fixed weight coefficients of positives and negatives. In this paper, we put effort on establish direct link between measurement and loss through variable weight coefficients.

%In object detection, the number of negative classes in an image is much more than the positive class. Since most of the negatives are more easy to be classified, mining difficult examples can reduce the weight of negative classes in the classification. However, in mismatch removal tasks, the more outliers, the more difficult it is to classify outliers. Therefore, mining difficult examples can not solve the problem of category imbalance. Besides, there is no method that explores how to establish a direct connection between Fn-score and loss function.

\noindent \textbf{Attention Mechanism} Attention mechanism focuses on perceiving salient areas similar to human visual systems. Non-local neural network \cite{wang2018non} adopts non-local operation to introduce attention mechanism in feature map. SE-Net \cite{hu2018squeeze} introduces channel-wise attention mechanism through a Squeeze-and-Excitation block. In order to explore second-order statistics, SAN-Net \cite{dai2019second} utilizes second-order channel attention (SOCA) operations in their network. In addition to the two dimensional convolution, Wang et. al propose a graph attention convolution (GAC) \cite{wang2019graph} for dealing with point cloud data. The above literature shows that attention mechanism can enhance the network performance of different tasks. In this paper, we are committed to design an attention mechanism block for mismatch removal.

%In the above methods, attention mechanism is introduced for perception, which means that you can improve network performance by amplifying salient features. However, mismacth removal is a fitting task. Attention mechanism in this task is to pay more attention in inliers during fitting, not just amplifying salient features.

%All papers submitted for publication by AAAI Press must be accompanied by a valid signed copyright form. There are no exceptions to this requirement. You must send us the original version of this form. However, to meet the deadline, you may fax (1-650-321-4457) or scan and e-mail the form (pubforms19@aaai.org) to AAAI by the submission deadline, and then mail the original via postal mail to the AAAI office. If you fail to send in a signed copyright or permission form, we will be unable to publish your paper. There are \textbf{no exceptions} to this policy.You will find PDF versions of the AAAI copyright and permission to distribute forms in the AAAI AuthorKit.

\begin{figure}[t]
	\centering
	\includegraphics[width=1\columnwidth]{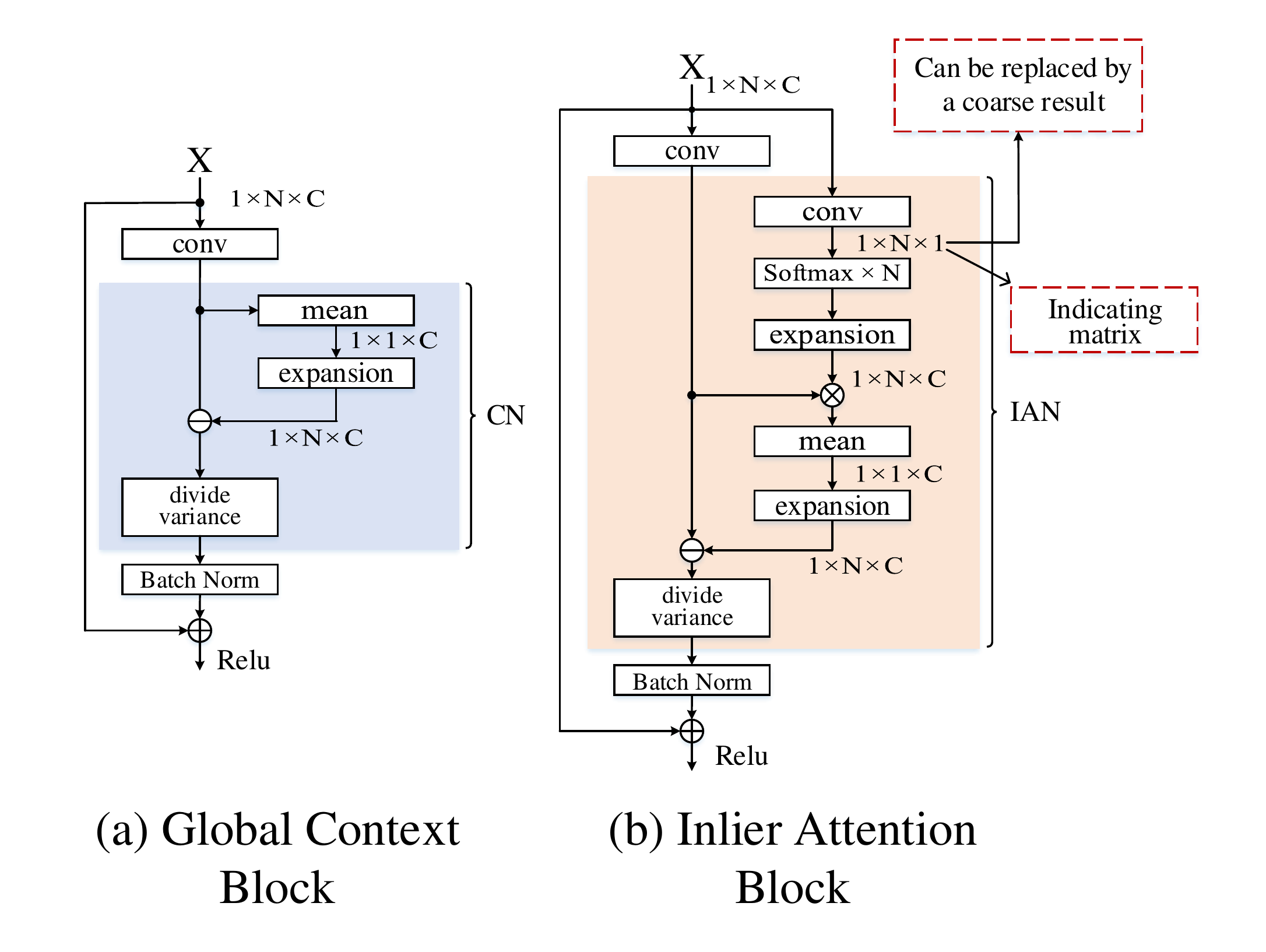} % Reduce the figure size so that it is slightly narrower than the column. Don't use precise values for figure width.This setup will avoid overfull boxes. 
	\caption{Comparison of Global Context Block and our IA Block.} 
	\label{base_architecture}
\end{figure}

\begin{figure}[t]
	\centering
	\includegraphics[width=1\columnwidth]{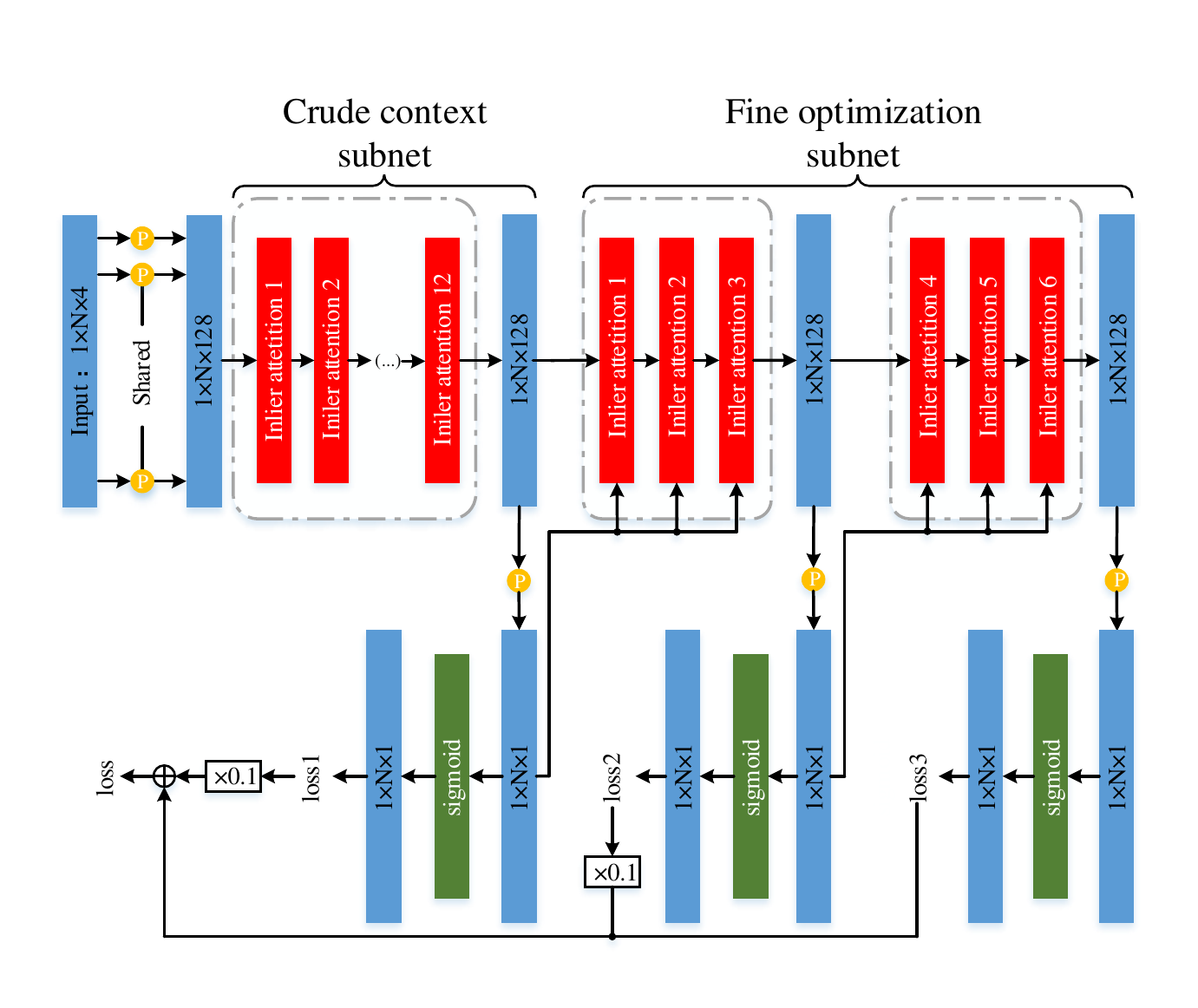} % Reduce the figure size so that it is slightly narrower than the column. Don't use precise values for figure width.This setup will avoid overfull boxes. 
	\caption{\textbf{GLA-Net Architecture}. The architecture of our GLA-Net consists of two subnet: the crude context subnet and fine optimization subnet. It is a process from coarse to fine.} 
	\label{wholeArchitecture}
\end{figure}

\section{Proposed Method}
%Our goal is to design and train a network which can meet two requirements: (1) A network that can fit a model by using the correct pairs of putative match pair. (2) A appropriate loss function that can guide the network to find the classification hyperplane that meets the specific requirements. The details of the proposed approach are discussed below.

\subsection{Problem Statement}
Suppose we have a pair of images $\mathcal{(I, I^{'})}$ 
with the coordinates of a putative correspondence set $\mathcal{(K=}$
$\{u,v\}$
$\mathcal{,K^{'}=}$
$\{u^{'},v^{'}\}$$\mathcal{)}$ between them. The putative set is obtained by performing nearest neighbor matching on handcrafted image descriptors (e.g., SIFT) or deep learning based image descriptors (e.g. LIFT). Our method formulates the mismatch removal task as a binary classification problem and determines whether a putative match is inlier or outlier.

\subsection{Guided Loss}
In the classification networks, the classification of positive and negative classes is treated as a single task. They usually utilize a cost sensitive loss to address imbalance between categories as follows:
\begin{equation} \label{original_loss}
loss=-\frac{1}{N}(\alpha\sum_{i=1}^{N_{pos}}\log(y_{i}) + \beta\sum_{j=1}^{N_{neg}}\log(1-y_{j})),
\end{equation}
where $N$ is the number of putative correspondences, and $N_{pos}$ and $N_{neg}$ are number of positive and negative instances respectively. $y_{i}$ and $y_{j}$ are the network output of instance $i$ and $j$. $\alpha$ and $\beta$ are the weight coefficients of positive and negative instances. We suggest inspecting this task from another point of view. Specifically, the classification of positives and negatives are regarded as two separate issues. Then, the loss function is reconstructed as follows:
\begin{equation}
\begin{aligned}\label{rewritten_loss}
loss=-(\lambda \frac{1}{N_{pos}} \sum_{i=1}^{N_{pos}}&\log(y_{i}) + \mu  \frac{1}{N_{neg}} \sum_{j=1}^{N_{neg}}\log(1-y_{j})), \\
& s.t. \quad \ \lambda + \mu = 1,
\end{aligned}
\end{equation}
where $\lambda$ and $\mu$ are also weight coefficients. The other variables have the same meaning as Eq. \ref{original_loss}. In machine learning theory, the logarithmic loss function and the 0-1 loss function have similar properties. In order to facilitate our subsequent derivation, we first replace the logarithmic loss with 0-1 loss. Suppose that after passing through the classifier, the number of misclassified instances in the positive and negative classes is $X$ and $Y$ respectively. Then the Eq. \ref{rewritten_loss} can be rewritten as:
\begin{equation}\label{simplify_loss}
loss=\lambda \frac{X}{N_{pos}}  + \mu \frac{Y}{N_{neg}}.
\end{equation}
Since positive and negative categories are treated as two independent tasks, X and Y are independent variables. 

Meanwhile, the precision ($P$) and recall ($R$) of the network can be calculated by $X$ and $Y$, and Fn-score ($Fn$) can be calculated by precision and recall. We present the relationships between these variables as follows:
\begin{equation}\label{Fn}
Fn = \dfrac{(1 + n^{2}) \cdot P \cdot R}{n^{2} \cdot P + R},
\end{equation}

\begin{equation}\label{PR}
P = \dfrac{N_{pos} -X}{N_{pos} - X + Y},R = \dfrac{N_{pos} -X}{N_{pos}}.
\end{equation}

Our motivation is to establish a direct connection between loss and Fn-score. Specifically, we hope that the decrease of the loss during training will definitely lead to an increase in Fn-score. From a mathematical point of view, that means Fn-score and loss are perfectly negatively correlated. This relationship can be expressed in the form of differential as follows:
\begin{equation}\label{dldf}
dloss \cdot dFn \leq  0.
\end{equation}
For the convenience of derivation, we sign $\dfrac{\partial loss}{\partial X}$ as $\partial l_{X}$ and $\dfrac{\partial Fn}{\partial X}$ as $\partial F_{X}$.
$Y$-related expressions are treated the same way as $X$. Then:
%\begin{equation}\label{dloss}
%\begin{aligned}
%dloss = \frac{\partial loss}{\partial X} dX + \frac{\partial loss}{\partial Y} dY, dFn = \frac{\partial Fn}{\partial X} dX + \frac{\partial Fn}{\partial Y} dY
%\end{aligned}
%\end{equation}

\begin{equation}\label{detail}
\begin{aligned}
dloss = \partial l_{X} dX + \partial l_{Y} dY, dFn = \partial F_{X} dX + \partial F_{Y} dY.
\end{aligned}
\end{equation}
As long as Eq. \ref{dldf} is true, the decline of loss will certainly lead to the rise of Fn-score. So we start our Guided Loss from Eq. \ref{dldf}. We substitute Eq. \ref{detail} into Eq. \ref{dldf}, then:

%\begin{equation}\label{constraint2}
%\begin{aligned}
%\dfrac{\partial loss}{\partial X} \cdot \dfrac{\partial Fn}{\partial X} \cdot (dX)^{2} + \dfrac{\partial loss}{\partial Y} \cdot \dfrac{\partial Fn}{\partial Y} \cdot (dY)^{2} + \\
%(\dfrac{\partial loss}{\partial X} \cdot \dfrac{\partial Fn}{\partial X} + 
%\dfrac{\partial loss}{\partial X} \cdot \dfrac{\partial Fn}{\partial X} ) \cdot dX \cdot dY  \leq 0
%\end{aligned}
%\end{equation}

\begin{equation}\label{constraint2}
\begin{aligned}
\partial l_{X} \cdot \partial F_{X} \cdot (dX)^{2} + \partial l_{Y} \cdot \partial F_{Y} \cdot (dY)^{2}  + \\
(\partial l_{X} \cdot \partial F_{Y} + \partial l_{Y} \cdot \partial F_{X}) \cdot dX \cdot dY  \leq 0.
\end{aligned}
\end{equation}
Since $X \geq 0, Y \geq 0$ during training, We can derive the following inequality from Eq. \ref{simplify_loss}, Eq. \ref{Fn} and Eq. \ref{PR}. 

%\begin{equation}\label{dloss}
%\begin{aligned}
%\dfrac{\partial loss}{\partial X}\geq 0 ,\dfrac{\partial loss}{\partial Y} \geq 0,  \dfrac{\partial Fn}{\partial X}\leq 0,\dfrac{\partial Fn}{\partial Y}\leq 0
%\end{aligned}
%\end{equation}

\begin{equation}\label{dloss}
\begin{aligned}
\partial l_{X}\geq 0 ,\partial l_{Y} \geq 0, \partial F_{X}\leq 0,\partial F_{Y}\leq 0.
\end{aligned}
\end{equation}
In order to express Eq. \ref{constraint2} more clearly, we transform it into the following form:

\begin{equation}\label{constraint3}
\begin{aligned}
(\sqrt{-\partial l_{X} \cdot \partial F_{X}} \cdot dX + \sqrt{-\partial l_{Y} \cdot \partial F_{Y}}\cdot dY)^{2} + \\
(\sqrt{-\partial l_{X} \cdot \partial F_{Y}} - \sqrt{-\partial l_{Y} \cdot \partial F_{X}})^{2} \cdot dX \cdot dY \geq 0.
\end{aligned}
\end{equation}
In order for Eq. \ref{constraint3} to be the permanent establishment, then
\begin{equation}\label{constraint4}
\begin{aligned}
\sqrt{-\partial l_{X} \cdot \partial F_{Y}} - \sqrt{-\partial l_{Y} \cdot \partial F_{X}} = 0,
\end{aligned}
\end{equation}
which is:
\begin{equation}\label{st}
\dfrac{\partial F_{X}}{\partial F_{Y}} = \dfrac{\partial l_{X}}{\partial l_{Y}}.
\end{equation}
Thus, Fn-score and loss are perfectly negatively correlated as long as the constraint of Eq. \ref{st} is satisfied during the training. Based on the above derivations, we present the specific flow of our Guided Loss algorithm as Algorithm \ref{alg:G_Loss}. Specifically, for each image pair in current training batch, we calculate the derivatives of Fn-score with respect to $X$ and $Y$ under the current network parameters. Then we can update $\lambda$ and $\mu$ by the constraint on Eq. \ref{st}. Each iteration will update $\lambda$ and $\mu$ before calculating the loss.

%It updates the $\lambda$ and $\mu$ through numerical derivation.
%So the coefficients of $(dX)^{2}$ and $(dY)^{2}$ in Eq. \ref{constraint2} are both negative. Therefore, Eq. \ref{constraint2} holds as long as it is perfect square trinomial. We set Eq. \ref{constraint2} to perfect square trinomial, which leads to the following conclusion:

%Then we can update $\lambda$ and $\mu$ with a numerical derivative method in each batch as Algorithm \ref{alg:G_Loss}.

\begin{algorithm} [h]%算法开始 
	\caption{Guided Loss} %算法的题目 
	\label{alg:G_Loss} %算法的标签 
	%	\algsetup{linenosize=\footnotesize}
	%	\footnotesize
	{\bf Input:} a batch of training data; last network parameter \\
	{\bf Output:} Proportion of positive and negative loss ($\lambda, \mu$)
	\begin{algorithmic}[1] %此处的[1]控制一下算法中的每句前面都有标号 

		\FOR{$i = 0; iter < Batch\_size; i ++$}
		\STATE Calculate $P_{i}$, $R_{i}$, $Fn_{i}$ of this image pair
		
		\STATE $\delta_{X_{i}} = - 1$, $\delta_{Y_{i}} = 0$ $\rightarrow$ $\delta_{Fn_{i}}$ $\rightarrow$ $\dfrac{\partial Fn_{i}}{\partial X_{i}} = -\delta_{Fn_{i}}$
		
		\STATE $\delta_{Fn^{'}_{i}} = \delta_{Fn_{i}}$, $\delta_{X^{'}_{i}} = 0$, $\rightarrow$ $\delta_{Y^{'}_{i}}$ $\rightarrow$ $\dfrac{\partial Fn_{i}}{\partial Y_{i}} = \dfrac{\delta_{Fn^{'}_{i}}}{\delta_{Y^{'}_{i}}}$ $\rightarrow$ $\dfrac{\partial Fn_{i}}{\partial X_{i}} / \dfrac{\partial Fn_{i}}{\partial Y_{i}} = -\delta_{Y_{i}^{'}}$
		
		\STATE compute $\dfrac{\partial loss_{i}}{\partial X_{i}}$ and $\dfrac{\partial loss_{i}}{\partial Y_{i}}$ according to Eq. \ref{simplify_loss} $\rightarrow$ $\dfrac{\partial loss_{i}}{\partial X_{i}}$ = $\dfrac{\lambda_{i}}{N_{pos_{i}}}$, $\dfrac{\partial loss_{i}}{\partial Y_{i}}$ = $\dfrac{\mu_{i}}{N_{neg_{i}}}$
		
		\STATE s.t. $\lambda_{i} + \mu_{i} = 1$ $\rightarrow$ compute $\lambda_{i}$ and $\mu_{i}$ according to Eq. \ref{st} and the results of step 4 and 5
		\ENDFOR
		\STATE return $\lambda$, $\mu$
	\end{algorithmic} 
\end{algorithm}

%In order to illustrate that our algorithm can make sure that Fn-score and loss are perfectly negatively correlated. We first list the differential forms of Fn-score and loss as follows:
%
%
%Then we can prove that as long as Eq. \ref{st} is true, then the following formula must be established:
%
%A detailed proof process can be available in Appendix A. Thus, the sign of the differential of Fn-score and loss is always the opposite. In other word, Fn-score and loss are perfectly negatively correlated. 

\subsection{Inlier Attention Block}
%In essence, deep learning based methods are also model-fitting methods. They fit a model that can removes mismatches by using the putative matching set. 

Since the input of the network for mismatch removal is the unordered feature correspondences, the feature extraction blocks are required to be permutation-equivariant. Global Context Blocks are utilized as the backbone architecture in LFGC-Net \cite{moo2018learning} and its variations \cite{Ploetz:2018:NNN,zhao2019nm}. The architecture of this block in as shown in Fig. \ref{base_architecture} (a). It extracts global context through a Context Normalization (CN) operation. CN is a simple operation that normalizes the feature maps through subtracting their means and dividing by their variances. It is very effective for processing unordered data. 

However, CN completely ignores the influence of outliers on model fitting, while the features of outliers may impair the global context. In order to mitigate the negative impact of outlier on the network, we propose a IA Block architecture as shown in Fig. \ref{base_architecture} (b). IA block replaces CN with an Inlier Attention Normalization (IAN) operation to introduce attention mechanism. Specifically, our IAN learns a soft outlier-free matrix through global context in each IA Block. This indicating matrix provides spatial variability for each instance. In SE-Net \cite{hu2018squeeze}, the indicating matrix is directly multiplied by the feature map to magnify salient features. We also leverage simple multiplication to introduce spatial differences.

Formally, let $o_{i}^{l} \in \mathbb{R}^{C^{l}}$ be the output of $i$-th correspondence in layer $l$, where $C^{i}$ is the channel number of layer $l$. Then CN and IAN operation can both be expressed as follows:

\begin{equation}
Output(o_{i}^{l}) = \frac{o_{i}^{l} - \mu^{l}}{\sigma^{l}},
\end{equation}
where in CN :
\begin{equation}
\mu^{l} = \frac{1}{N}\sum_{i=1}^{N}o_{i}^{l},\sigma^{l} = \sqrt{\frac{1}{N}\sum_{i=1}^{N}(o_{i}^{l}-\mu^{l})^2}.
\end{equation}
IAN calculates $\sigma^{l}$ in the same way as CN, but obtains $u^{l}$ in a different way, as follows:
%\begin{equation}
%\begin{aligned}
%\mu^{l} = \frac{1}{N}\sum_{i=1}^{N}((sigmoid(r^{l}) * N) .* o_{i}^{l}) \\
%r^{l} = conv(o_{i}^{l})
%end{aligned}
%\end{equation}

\begin{equation}
\begin{aligned}\label{eq2}
\mu^{l} = \frac{1}{N}\sum_{i=1}^{N}((soft&max(r^{l}) * N) .* o_{i}^{l}), \\
r^{l} = conv&(o_{i}^{l}),
\end{aligned}
\end{equation}
where $r^{l}$ is the indicating matrix in each IA Block. Since CN integrates global context mainly by the operation of subtracting mean, we use the indicating matrix to participate in the calculation of the mean, instead of changing the feature map. This change preserves the feature of each correspondence and filters outlier information through a weighted mean calculation.
It down-weights the impacts of outliers in the process of global context extraction. 
Moreover, the indicating matrix can not only be automatically learned over the network, but can also replaced directly by a preliminary classification result. The visual comparison between Global Context Block and IA Block (CN and IAN operations are highlighted) is shown in the Fig. \ref{base_architecture}.
%\begin{figure}[t]
%	\centering
%	\subfigure[Curve diagram of St.Brown dataset]
%	{
%		\begin{minipage}[b]{0.5\textwidth}
%			\includegraphics[width=1\textwidth]{fig//trainingPR2.pdf}
%		\end{minipage}
%	}
%	\subfigure[Curve diagram of COLMAP dataset]
%	{
%		\begin{minipage}[b]{0.5\textwidth}
%			\includegraphics[width=1\textwidth]{fig//trainingPR3.pdf}
%		\end{minipage}
%	}
%	\caption{Precision and recall of the networks with different losses in each training iteration. L-Loss is the loss function that LFGC-Net \cite{moo2018learning} used. It's a cost sensitive loss with fixed weight coefficients. G-Loss is the proposed loss with F1-score's guidance. FocalLoss \cite{lin2017focal} is a commonly used loss in object detection.} \label{fig:trainingPR}
%\end{figure}

\begin{figure}[t]
	\centering
	\includegraphics[width=1\columnwidth]{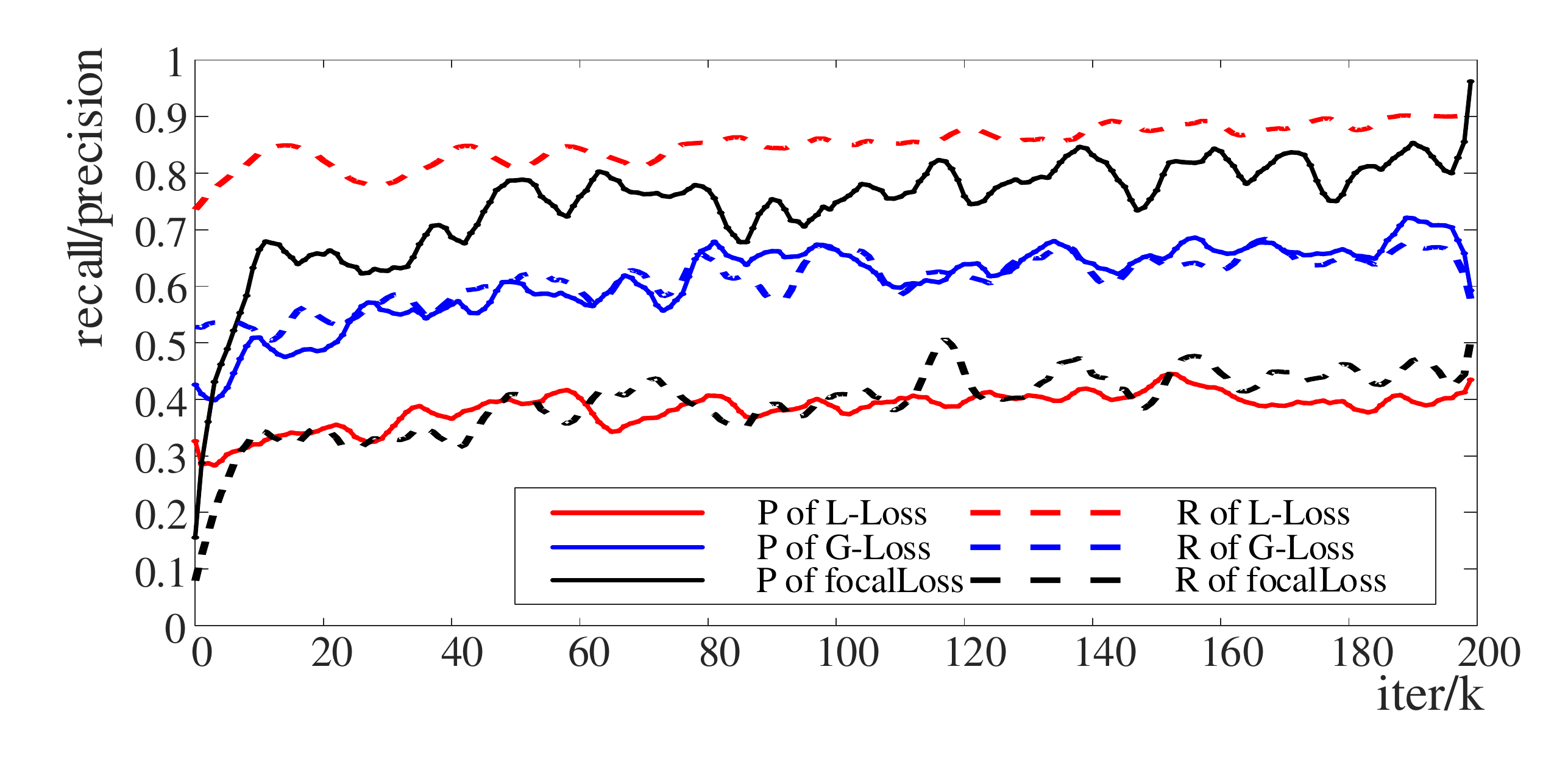} % Reduce the figure size so that it is slightly narrower than the column. Don't use precise values for figure width.This setup will avoid overfull boxes. 
	\caption{Precision and recall of the networks with different losses in each training iteration on the St.Brown dataset. L-Loss is the loss function that LFGC-Net \cite{moo2018learning} used. It is a cost sensitive loss with fixed weight coefficients. G-Loss is the proposed loss with F1-score's guidance. FocalLoss \cite{lin2017focal} is a commonly used loss in object detection.} \label{fig:trainingPR}
\end{figure}

%In the networks of two-dimensional image data, attention mechanism can be divided into spatial attention (e.g. Non-local network, \cite{wang2018non}) and channel-wise attention (e.g. SE-Net, \cite{hu2018squeeze}). From this classification method, IA Block is a method of spatial attention. What's interesting, however, is that our method just extracts global context for computing the soft outlier-free matrices, which is more similar to channel-wise attention in two-dimensional image processing. As shown in Fig. \ref{base_architecture} (b), IAN is followed by a Batch normalization and a Relu activation function in each IA Block.

\begin{table}[t]%%If you do not follow these requirements, your paper will be subject to expensive reformatting and special handling fees that can easily exceed the extra page fee.
	\centering%
	\caption{The datasets we use to evaluate the performances of different networks. VP means viewpoint.} \label{Dataset}%
	
	\begin{tabular}{c|c|c|c}%\item The \LaTeX{}-generated files (e.g. .aux and .bbl file, etc.) if they are needed to compile your source (generally, the .aux and bbl file can be omitted; the bib file might be necessary, however).
		
		\hline%\end{itemize}
		%
		%Your \LaTeX{} source will be reviewed and recompiled on our system (if it does not compile, you may incur fees). \textbf{Do not submit your source in multiple text files.} Your single \LaTeX{} source file must include all your text, your bibliography (formatted using aaai.bst), and any custom macros. 
		\textbf{Dataset} & {\textbf{\tabincell{c}{\# Image\\
					Pair}}} & {\textbf{\tabincell{c}{\# Inlier\\
					Ratio(\%)}}}  & \textbf{Chanllenges}\\ \hline%
		
		St.Brown & 16170 & 7.59 & {\tabincell{c}{Scenario \\ changes}}   \\ \hline
		WIDE & 11426 & 32.77 & {VP changes}   \\ \hline 
		COLMAP & 18850 & 7.50 & {\tabincell{c}{VP changes\\
				\& rotation}} \\ \hline
	\end{tabular}%\usepackage{aaai19}
\end{table}

%\begin{figure}[t]
%	\centering
%	\includegraphics[width=1\columnwidth]{fig//trainingPR2.pdf} % Reduce the figure size so that it is slightly narrower than the column. Don't use precise values for figure width.This setup will avoid overfull boxes. 
%	\caption{Precision and recall on training data of different loss during training . L-Loss is the loss function that LFGC-Net used. It's a cost sensitive loss with fixed weight coefficients. G-Loss is our proposed loss with F1-score's guidance. FocalLoss \cite{lin2017focal} is a commonly used loss in object detection.} 
%	\label{introduction}
%\end{figure}

\begin{figure}[t]
	\centering
	\includegraphics[width=1\columnwidth]{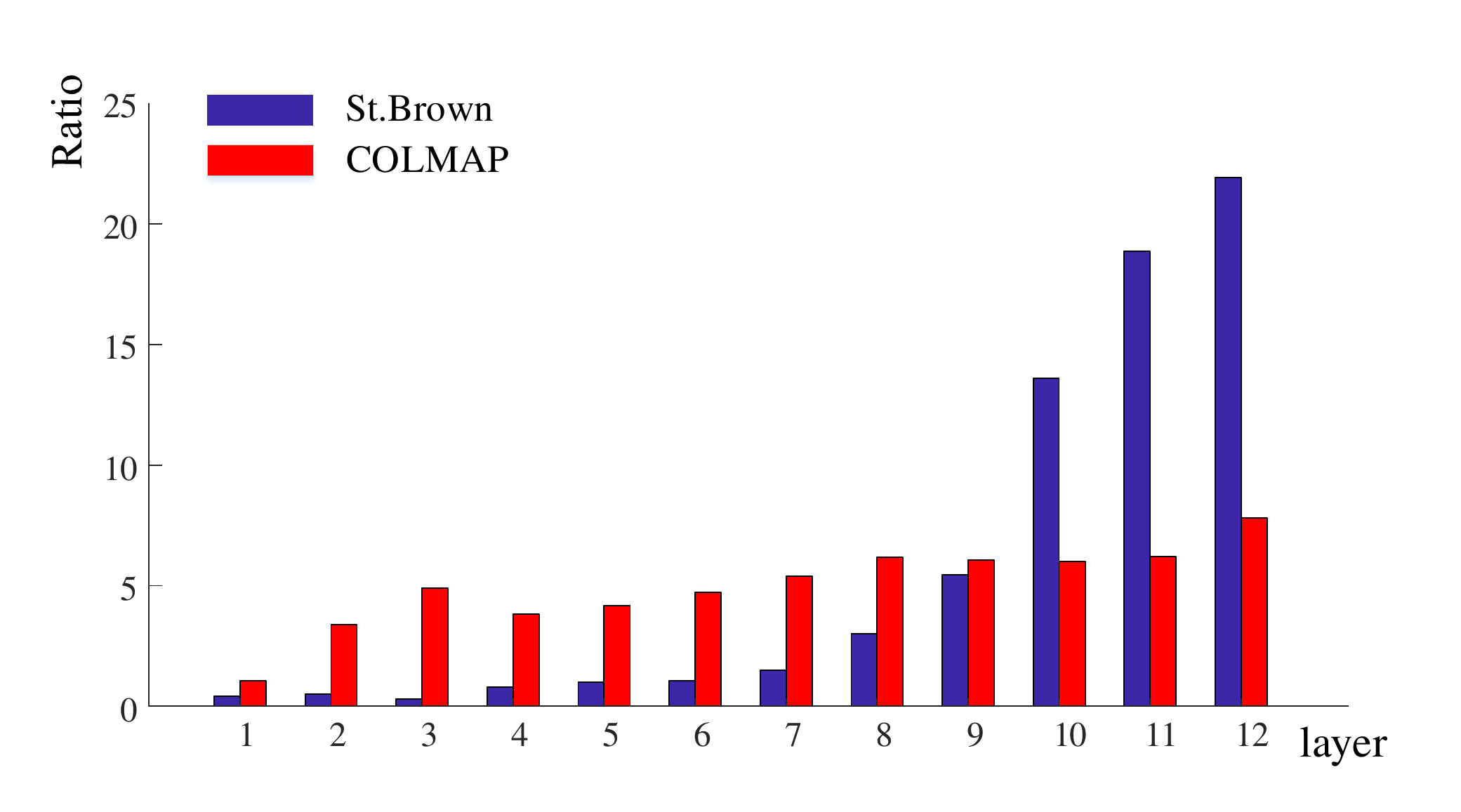} % Reduce the figure size so that it is slightly narrower than the column. Don't use precise values for figure width.This setup will avoid overfull boxes. 
	\caption{Attention mechanism analysis. The average ratio of weights assigned to inliers and outliers by the indicating matrix are calculated on St.Brown and COLMAP dataset to examine if the indicating matrix can assign higher weights to inliers.} 
	\label{bar}
\end{figure}

\subsection{GLA-Net}
The whole architecture of GLA-Net is shown as Fig. \ref{wholeArchitecture}. Inspired by the local optimization operations in geometric model-fitting methods \cite{chum2003locally,barath2018graph}, the GLA-Net is developed in a coarse-to-fine manner.
% Our network is partially inspired by LFGC-Net \cite{moo2018learning} and LO-RANSAC \cite{chum2003locally}. In general, GLA is a process from coarse to fine.
As shown in Fig. \ref{wholeArchitecture}, the network can be divided into two sub-networks: crude context subnet and fine optimization subnet. Crude context subnet integrates the global context with our IA Block to get a preliminary result. During training, the preliminary result is supervised by the auxiliary loss ($loss1$ and $loss2$ in Fig. \ref{wholeArchitecture}), which is a commonly used trick in many networks, such as GoogLeNet \cite{szegedy2015going}. 
Fine optimization subnet is designed to perform local optimization to achieve a better result. Since preliminary result has been obtained in previous subnet, IA Blocks do not learn indicating matrix, but treat the preliminary result as indicating matrix instead. The same optimization process is performed twice in this subnet.

%It also utilizes the IA Block to extract features, but the indicator matrices in this subnet are the supervised results from previous layers. 

%The fine optimization subnet contains two identical structures, and the final classification result is output by the last layer of the network.

To unite the two subnets, we combine the losses of intermediate and final results as total loss:
\begin{equation}
\begin{aligned} \label{wholeLoss}
loss=\rho loss1+\eta loss2+loss3, 0 \leq \rho \leq 1, 0 \leq \eta \leq 1,
\end{aligned}
\end{equation}
where $loss3$ is the main loss, and $loss1$ and $loss2$ are the auxliary losses. $\rho$ and $\eta$ are used to adjust the weights of auxliary losses.

% since $loss1$ and $loss2$ are to monitor the network to get preliminary result, we calculate them cost sensitive loss with fixed weights of positives and negatives. The weights of positive and negative classes are determined their number. $loss3$ is used to monitor the final classification result, so $loss3$ is obtained by our Guided Loss.

%The coarse step of our network is similar to LFGC-Net \cite{moo2018learning}. We use the ResNet Block with context normlization and batch normlization as the basic block to extract feature. After perferming several ResNet Blocks, the unordeered raw correspondences are converted to some feature vectors. These features are used both as input to subsequent networks and to guide a coarse result with a auxiliary loss. The fine step consists of two identical optimized structures, each of which leads to a more refined result. In this step, we adopt our Inlier Attention Convolution (IAC) Block as the basic block. As mentioned before, IAC block requires a feature vector and a coarse classification result as input. Each IAC block takes the features of last layer and last coarse result as input. During training, the three losses are combined to optimize the network parameters, and the total loss is as follows：
%
%
%where $loss3$ is the main loss, while $loss1$ and $loss2$ are the auxliary losses.

\section{Experiments}
In this section, we first evaluate the Guided Loss and IA Block separately, and then we evaluate the overall performance with the state of the art methods. All experiments are run on Ubuntun 16.04 with NVIDIA GTX1080Ti.

\subsection{Experimental Setup}
\noindent\textbf{Parameter Settings} In our network, all the convolution kernels' size is $(1 \times 1)$. In Eq. \ref{wholeLoss}, $\rho$ and $\eta$ are both set to 0.1. The auxiliary losses ($loss1$ and $loss2$) are calculated by L-Loss and the main loss ($loss3$) is by our Guided Loss, where L-Loss is the loss function used in LFGC-Net \cite{moo2018learning}. Our GLA-Net is trained by Adam with a learning rate being $10^{-3}$ and batch size being 16.

\begin{table}[t]%%If you do not follow these requirements, your paper will be subject to expensive reformatting and special handling fees that can easily exceed the extra page fee.
	\centering%
	\caption{Performances of different loss functions on St.Brown and COLMAP datasets. L-Loss is the loss function that LFGC-Net \cite{moo2018learning} used. FocalLoss \cite{lin2017focal} is a commonly used loss in object detection. F1-Loss, F2-Loss and F0.5-Loss are the Guided Losses with the guidances of F1-score, F2-score and F0.5-score respectively.} \label{tab:guideLoss}%
	\footnotesize
	\begin{tabular}{c|c|c|c|c|c}%\item The \LaTeX{}-generated files (e.g. .aux and .bbl file, etc.) if they are needed to compile your source (generally, the .aux and bbl file can be omitted; the bib file might be necessary, however).
		
		\hline%\end{itemize}
		%
		%Your \LaTeX{} source will be reviewed and recompiled on our system (if it does not compile, you may incur fees). \textbf{Do not submit your source in multiple text files.} Your single \LaTeX{} source file must include all your text, your bibliography (formatted using aaai.bst), and any custom macros. 
		\textbf{Method} & \textbf{P(\%)} & \textbf{R(\%)} & \textbf{F1(\%)}  & {\textbf{F2(\%)}} & {\textbf{F0.5(\%)}} \\ \hline%
		\multicolumn{6}{c}{Dataset : St.Brown} \\ \hline
		
		L-Loss & 39.72 & \textbf{76.83} & 49.83 & {61.35} & {43.08}  \\ \hline 
		Focal-Loss & \textbf{70.67} & 41.44 & {49.67} & {52.77} & \textbf{69.11} \\ \hline 
		F1-Loss & 58.29 & 56.04 &  \textbf{56.32} &{57.17} & {58.67}  \\ \hline 
		F2-Loss & 47.62 & 66.20 & {54.24} & \textbf{62.05}  & {51.01} \\ \hline
		F0.5-Loss & 64.64 & 51.11 & {56.02} & {54.66} & {63.57} \\ \hline \hline 
		
		\multicolumn{6}{c}{Dataset : COLMAP} \\ \hline
		
		L-Loss & 27.95 & \textbf{65.63} & 34.83 & {45.67} & {31.03} \\ \hline 
		Focal-Loss & 27.52 & 49.44 & {32.32} & {42.03} & 31.78\\ \hline 
		F1-Loss & 33.96 & 48.62 & \textbf{38.10} & {45.68} & {37.96} \\ \hline 
		F2-Loss & 32.01 & 49.68 & 37.57 & \textbf{48.26} & 37.80  \\ \hline
		F0.5-Loss & \textbf{38.01} & 40.96 & {37.12} & {40.03} & \textbf{40.06} \\ \hline

	\end{tabular}%\usepackage{aaai19}
\end{table}%\usepackage{times}

\begin{table}[t]%%If you do not follow these requirements, your paper will be subject to expensive reformatting and special handling fees that can easily exceed the extra page fee.
	\centering%
	\caption{The classification results of LFGC-Net with the same loss function  but different backbones on COLMAP and WIDE datasets.} \label{tab:IAC}%
	\footnotesize
	\begin{tabular}{c|c|c|c}%\item The \LaTeX{}-generated files (e.g. .aux and .bbl file, etc.) if they are needed to compile your source (generally, the .aux and bbl file can be omitted; the bib file might be necessary, however).
		
		\hline%\end{itemize}
		%
		%Your \LaTeX{} source will be reviewed and recompiled on our system (if it does not compile, you may incur fees). \textbf{Do not submit your source in multiple text files.} Your single \LaTeX{} source file must include all your text, your bibliography (formatted using aaai.bst), and any custom macros. 
		\textbf{Method} & \textbf{P(\%)} & \textbf{R(\%)} & \textbf{F1(\%)} \\ \hline%
%		\multicolumn{4}{c}{Dataset : st.brown} \\ \hline
		
%		LFGC & \textbf{58.29} & 56.04 &  {56.32}  \\ \hline
%		LFGC with NM-Net-sp & 55.02 & {57.71} & {55.50} \\ \hline 
%		LFGC with SE-Block & 36.19 & 59.03 & {42.98} \\ \hline 
%		LFGC with IA-Block & 58.01 & \textbf{64.33} & \textbf{59.95} \\ \hline
		
		\multicolumn{4}{c}{Dataset : COLMAP} \\ \hline
		
		LFGC &  {33.96} & 48.62 & {38.10}  \\ \hline 
		LFGC with NM-Net-sp & 35.42 & {47.12} & {38.25} \\ \hline
		LFGC with SE-Block & 32.81 & 44.01 & {35.98} \\ \hline 
		LFGC with IA-Block & \textbf{37.28} & \textbf{62.08} &  \textbf{43.20} \\ \hline \hline
		
		\multicolumn{4}{c}{Dataset : WIDE} \\ \hline 
		
		LFGC & 93.28 & {94.44} & 93.62  \\ \hline 
		LFGC with NM-Net-sp & 92.74 & 92.83 & {92.75} \\ \hline
		LFGC with SE-Block & 93.58 & 93.28 & {93.28} \\ \hline 
		LFGC with IA-Block & \textbf{94.25} & \textbf{94.52} &  \textbf{94.30} \\ \hline

	\end{tabular}%\usepackage{aaai19}
\end{table}%\usepackage{times}

\begin{table}[t]%%If you do not follow these requirements, your paper will be subject to expensive reformatting and special handling fees that can easily exceed the extra page fee.
	\centering%
	\caption{Classification results of all ablation experiments on COLMAP dataset.} \label{tab:Ablation}%
	\footnotesize
	\begin{tabular}{c|c|c|c}%\item The \LaTeX{}-generated files (e.g. .aux and .bbl file, etc.) if they are needed to compile your source (generally, the .aux and bbl file can be omitted; the bib file might be necessary, however).
		
		\hline%\end{itemize}
		%
		%Your \LaTeX{} source will be reviewed and recompiled on our system (if it does not compile, you may incur fees). \textbf{Do not submit your source in multiple text files.} Your single \LaTeX{} source file must include all your text, your bibliography (formatted using aaai.bst), and any custom macros. 
		 & \textbf{P(\%)} & \textbf{R(\%)} & \textbf{F1(\%)} \\ \hline%
		%		\multicolumn{4}{c}{Dataset : St.Brown} \\ \hline
		
		%		LFGC & \textbf{58.29} & 56.04 &  {56.32}  \\ \hline
		%		LFGC with NM-Net-sp & 55.02 & {57.71} & {55.50} \\ \hline 
		%		LFGC with SE-Block & 36.19 & 59.03 & {42.98} \\ \hline 
		%		LFGC with IA-Block & 58.01 & \textbf{64.33} & \textbf{59.95} \\ \hline

		Baseline &  {27.95} & 65.63 & {34.83}  \\ \hline 
		Guided Loss & 33.96 & {48.62} & {38.10} \\ \hline
		IA Block & 30.13 & \textbf{82.06} & {38.10} \\ \hline 
		Full Framework & \textbf{39.20} & 58.82 &  \textbf{44.12} \\ \hline

	\end{tabular}%\usepackage{aaai19}
\end{table}%\usepackage{times}

\noindent\textbf{Benchmark Dataets} All of our contrast experiments are conducted on three challenging benchmark datasets: St.Brown \cite{moo2018learning}, WIDE \cite{zhao2019nm} and COLMAP \cite{zhao2019nm}. For each dataset, their camera parameters and ground-truth labels are obtained by Structure from Motion \cite{wu2013towards}. The basic properties of these three datasets are shown in Tab. \ref{Dataset}. During training, they are divided into disjoint subsets for training (70\%), validation (15\%) and testing (15\%).

\noindent\textbf{Evaluation Criteria} To measure the mismatch removal performance, we employ precision (P), recall (R), Fn-score (Fn)
and average deviation between the essential matrix E estimated by selected correspondences and ground-truth $E_{gt}$ (MSE). The Fn-score can be computed as Eq. \ref{Fn}. In Fn, $n$ is used to adjust the importance of precision and recall in the measurement. All experimental measurements in this paper are the average values of all instances on the corresponding dataset.

%\noindent\textbf{Evaluation Criteria} To measure the mismatch removal performance, we employ precision (P), recall (R), Fn-score (F) and average deviation between the essential matrix
%$E$ estimated by selected correspondences and ground-truth $E_{gt}$ (MSE). The Fn-score can be computed as follows:
%\begin{equation}
%\begin{aligned} \label{wholeLoss}
%Fn = \dfrac{(1 + \beta^{2}) \cdot P \cdot R}{\beta^{2} \cdot P + R}
%\end{aligned}
%\end{equation}

%The fixed proportion of positive and negative loss will cause the network to focus too much on the correct rate of a certain type in the later stage of training, resulting in low accuracy or low recall rate. In the experiment, we use F1-score (pay attention to both accuracy and recall rate) to guide the proportion of positive and negative classes in the loss function. Therefore, Accuracy and recall rate can be balanced and synchronized during training.

%\definecolor{myGreen}{RGB}{40, 180, 99}
%\definecolor{myColor2}{RGB}{127, 140, 141}
%\definecolor{myColor3}{RGB}{52, 152, 219}  

\begin{table}[t]%%If you do not follow these requirements, your paper will be subject to expensive reformatting and special handling fees that can easily exceed the extra page fee.
	\centering%
	\caption{Evaluation results on COLMAP, WIDE and St.Brown datasets.} \label{tab:overallPreformance}%
	\footnotesize
%	\begin{tabular}{|c||>{\columncolor{myGreen} }c||>{\columncolor{myColor2} }c||>{\columncolor{myColor3} }c||c|@{}}%\item The \LaTeX{}-generated files (e.g. .aux and .bbl file, etc.) if they are needed to compile your source (generally, the .aux and bbl file can be omitted; the bib file might be necessary, however).
		
	\begin{tabular}{|c|c|c|c|c|}
		
		\hline%\end{itemize}
		%
		%Your \LaTeX{} source will be reviewed and recompiled on our system (if it does not compile, you may incur fees). \textbf{Do not submit your source in multiple text files.} Your single \LaTeX{} source file must include all your text, your bibliography (formatted using aaai.bst), and any custom macros. 
		\textbf{Method} & \textbf{P(\%)} & \textbf{R(\%)} & \textbf{F1(\%)} & \textbf{ MSE }  \\ \hline \hline%
		\multicolumn{5}{|c|}{Dataset : COLMAP} \\ \hline 
		
		RANSAC & 25.156 & {14.477} & 17.464 & 1.984 \\
		
		GC-RANSAC & 18.785 & {44.239} & 21.226 & 2.976 \\
		
		%		GTM & 22.931 & 19.913 & {19.075} & 2.004 & 3.073 & {3.245} & 5.102 & 4.804 \\ 
		LMR & \textbf{50.776} & 26.070 & 32.482 & \textbf{1.962} \\ 
		PointNet & 13.596 & 41.765 & 19.710 & 2.051\\ 
		%		PointNet++ & 18.659 & 41.953 & {24.301} & 2.060 & 2.902 & {3.200} & 4.877 & 4.150   \\ 
		LFGC-Net & 27.945 & \textbf{65.630} & {34.826} & 2.007 \\ 
		%		NM-Net-sp & 29.296 & 59.710 & {37.503} & 1.983 & 2.446 & {3.047} & 5.125 & 2.250   \\ 
		
		$\rm N^{3}$-Net & - & - & - & -\\
		
		NM-Net & 31.993 & 54.484 & {38.916} & 1.980 \\ 
		Ours & {39.200} & 58.819 & \textbf{44.124} & 1.985\\ \hline\hline
		
		\multicolumn{5}{|c|}{Dataset : WIDE} \\ \hline 
		
		RANSAC & 80.740 & {51.198} & 60.350 & 2.052  \\
		%		GTM & 79.989 & 47.711 & {58.881} & 2.040 & 2.784 & {3.068} & 5.041 & 1.046 \\ 
		GC-RANSAC & 68.097 & {88.766} & 76.151 & 2.151 \\
		
		LMR & 84.167 & 78.739 & 82.720 & 2.306\\ 
		PointNet & 64.730 & 77.287 & 70.068 & 2.282 \\
		%		PointNet++ & 73.926 & 81.856 & {77.245} & 2.180 & 2.771 & {3.255} & 5.013 & 3.020   \\ 
		LFGC-Net & 92.203 & 96.530 & {93.811} & 1.765 \\ 
		%		NM-Net-sp & 91.742 & 94.039 & {92.749} & 2.513 & 2.731 & {3.751} & 5.110 & 0.650   \\ 
		
		$\rm N^{3}$-Net & 86.931 & 96.402 & {90.889} & 3.030\\
		
		NM-Net & 93.280 & 94.752 & {93.922} & 2.430\\ 
		Ours & \textbf{94.469} & \textbf{96.627} & \textbf{95.214} & \textbf{1.731} \\ \hline\hline
		
		%		\multicolumn{9}{|c|}{Dataset : NARROW} \\ \hline
		%		
		%		RANSAC & 86.923 & \textbf{60.397} & 69.194 & 2.017 & \textbf{2.622} & 2.809 & \textbf{4.978} & 0.755  \\
		%		GTM & 88.707 &  {52.949} & 65.653 & 2.042 & {2.728} & 2.886 & 4.968 & 2.467 \\ 
		%		LMR & 0 & 0 & 0 & 0 & 0 & 0 & 0 & 0  \\ 
		%		PointNet & 79.003 & 86.163 & 82.102 & 2.293 & 2.787 & 3.503 & 4.728 & 1.180 \\
		%%		PointNet++ & 83.677 & 85.045 & {84.112} & 2.248 & 2.773 & {3.328} & 5.128 & 3.899   \\ 
		%		LGC-Net & 95.857 & 98.592 & {96.825} & 2.320 & 2.739 & {3.091} & 4.057 & 1.33   \\ 
		%%		NM-Net-sp & 96.946 & 97.659 & {97.283} & 2.482 & 2.664 & {3.687} & 5.038 & 0.245   \\
		%		NM-Net & 97.169 & 97.870 & {97.501} & 2.436 & 2.608 & {3.630} & 5.021 & 0.390   \\ 
		%		Ours & 97.776 & 98.332 & {98.025} & 2.383 & 2.789 & {3.117} & 4.006 & 1.036   \\ \hline %\hline
		
		\multicolumn{5}{|c|}{Dataset : St.Brown} \\ \hline
		
		RANSAC & 23.743 & {40.831} & 29.322 & 2.659\\
		% GTM & 88.707 &  {52.949} & 65.653 & 2.042 & {2.728} & 2.886 & 4.968 & 2.467 \\ 
		
		GC-RANSAC & 20.833 & {38.473} & 26.516 & 2.486\\
		
		LMR & 50.272 & 26.077 & 32.160 & \textbf{2.362}\\ 
		PointNet & 27.820 & 47.223 & 32.475 & 3.209 \\
		%		PointNet++ & 83.677 & 85.045 & {84.112} & 2.248 & 2.773 & {3.328} & 5.128 & 3.899   \\ 
		LFGC-Net & 39.72 & \textbf{76.83} & {49.83} & 3.042\\ 
		%		NM-Net-sp & 96.946 & 97.659 & {97.283} & 2.482 & 2.664 & {3.687} & 5.038 & 0.245   \\
		
		$\rm N^{3}$-Net & 40.923 & 75.341 & {51.683} & 3.023\\ 
		
		NM-Net-sp & 40.659 & 71.663 & {50.740} & 3.064\\ 
		Ours & \textbf{58.02} & 64.33 & \textbf{59.95} & 2.925 \\ \hline %\hline

	\end{tabular}%\usepackage{aaai19}
\end{table}%\usepackage{times}
\subsection{Submodule Performance}

As discussed in method description section, our method contains two main components including Guided Loss and IA Block.

\noindent\textbf{Guided Loss} As shown in Tab. \ref{tab:guideLoss}, to evaluate the performance of our Guided Loss, we train the same network architecture (In this experiments, LFGC-Net is used) with different loss functions, including L-Loss and Focal Loss. To illustrate that our Guided Loss can be employed for various measurements, we adopt different Fn-score to guide the network. In most cases, with the guidance of Fn-score, the network can achieve the best performance on this kind of measurement. This indicates that establishing a direct link between loss and measurement is conducive to improving network performance. Besides, different guiding indicators also show the stability of our guiding algorithm.

%This indicates that Guided Loss can guide the training process of the network more directly. Furthermore, different Fn-score  are suitable for different application scenarios. (F2-score pays more attention to recall. F0.5-score pays more attention to precision. F1-score tends to balanced precision and recall). Therefore, our Guided Loss can be applied to tasks with different requirements.
\begin{figure*}[t]
	\centering
	\includegraphics[width=1.8\columnwidth]{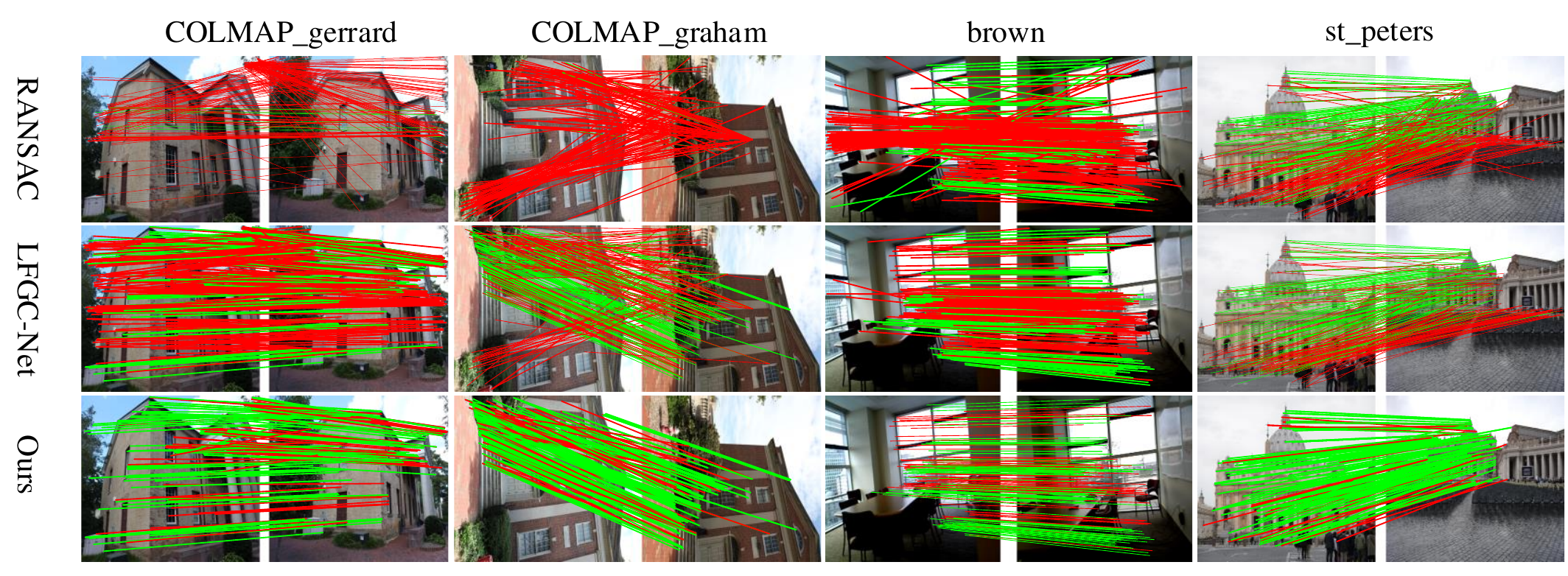} % Reduce the figure size so that it is slightly narrower than the column. Don't use precise values for figure width.This setup will avoid overfull boxes. 
	\caption{Visual comparison of matching results using RANSAC, LFGC-Net and our method. Images are taken from COLMAP and St.Brown datasets. Correspondences are in green if they are inliers, and in red otherwise. \textbf{Best viewed in color.}} 
	\label{visualize}
\end{figure*}
\noindent\textbf{Inlier Attention Block} We also show the effect of our IA Block from the results of precision, recall and F1-score measurement. To directly compare the performance of the backbone block, all comparison experiments utilize the LFGC-Net network framework with our F1-score Guided Loss, and only replace the backbone of the network.
As shown in Tab. \ref{tab:IAC}, LFGC is the original LFGC-Net, and NM-Net-sp is the backbone of NM-Net \cite{zhao2019nm} with spatial neighborhood mining. SE-Block is the attention mechanism backbone of SE-Net \cite{hu2018squeeze}. In order to make it more suitable for mismatch removal task, we changed it to the form of spatial attention. IA Block is our proposed architecture. 

As shown in Tab. \ref{tab:IAC}, our architecture has achieved leadership in almost all of the three evaluation metrics (P, R, F1). Therefore, the introduction of attention mechanism can improve the performances of mismatch removal networks. In addition, the comparison with SE-Net also shows that it is better to use the indicating matrix for model fitting instead of magnifying the feature map.

% when compared with LFGC and NM-Net-sp, which do not use attention mechanism, IA Block achieve better performance than them on precision, recall and F1-score. This indicates that introducing attention mechanism to mismatch removal network can improve network performance. When compared with SE-Block, IA Block can also have better performance than it. Thus, our IA Block is a suitable attention architecture for mismatch removal task. 

% In different application scenarios, mismatch removal task has different requirements. Some scenes need to retain more matching points, some scenes require higher precision, and some scenes require a balanced result. In order to adapt to different requirements, we guide the loss function with different F-scores. As shown in Tab. \ref{tab:guideLoss}, we list the precision (P), recall rate (R), F1-score (F1), F2-score (F2), F0.5-score (F0.5) of different losses. The network trained by LFGC and focal loss is not balanced in the precision and recall rate of the test set, so F-measure is not as good as ours. Besides, with the guidances of different F-measure, the network can achieve the best performance in this F-measure.

\subsection{Method Analysis}

In order to explore the working mechanism of each module, we analyze the training process and intermediate output of the network in detail, and conduct ablation studies.

\noindent\textbf{Training curve diagram} We record the precision and the recall of LFGC-Net with different losses on the training set during training in Fig. \ref{fig:trainingPR}. It can directly reflect the difference between the loss of the fixed and the variable weights. The L-Loss and focal loss all adopt a fixed weights of positive and negative classes, while G-Loss are with variable weights. The losses with fixed weights are more biased towards one measurement and ignores the other. It is difficult to assign proper weights to positive and negative classes by hand adjustment. Our G-loss can achieve a balance between precision and recall during training.

%L-Loss sets the weight of positive classes too large, which leads to a good network classification effect on positive classes and poor negative classes. Thus, LFGC holds high recall and low precision. Focal loss is the reverse. Our G-Loss adjust the ratios ($\lambda$, $\mu$) in real time during training. Therefore, Accuracy and recall of G-Loss can be balanced and synchronized during training.

\noindent\textbf{Indicating matrix}
In order to verify that our IA Block can be more biased to the inliers when fitting the model, we train a classification network with 12 IA blocks and record the average ratio of weights assigned to inliers and outliers. In order to confirm that the indicating matrices are automatically learned by the network, we do not use auxiliary loss trick in this experiment. As Fig. \ref{bar}, the average ratio remains above 1 in most layers and increases in the later layers of the network. It shows that our IA Block is more biased towards the context of inliers, thus filtering out the influence of outliers to some extent. And the attention effect will gradually increase as the number of network layers deepens. 

\noindent\textbf{Ablation study} We perform ablation study on the St.Brown dataset as shown in Tab. \ref{tab:Ablation}. The baseline network of our method is the LFGC-Net. We replace the feature extraction module and loss function in LFGC-Net with our IA Block and Guided Loss respectively. As shown in Tab. \ref{tab:Ablation}, the Guided Loss can achieve a balance between precision and recall, and IA Block can directly improve these two measurement. Combining these two modules can significantly improve network performance.

\subsection{The Overall Performance}

To test the effectiveness of our GLA-Net, we compare our GLA-Net with 7 state-of-the-art methods for mismatch removal: RANSAC \cite{fischler1981random}, GC-RANSAC \cite{barath2018graph}, LMR \cite{ma2019lmr}, pointNet \cite{qi2017pointnet}, LFGC-Net \cite{moo2018learning}, $ \rm N^{3}$-Net \cite{Ploetz:2018:NNN}, NM-Net \cite{zhao2019nm}. $\rm N^{3}$ cannot converge when training on the COLMAP dataset, so we do not put corresponding results on the table. NM-Net needs affine information to mine the neighborhood, while St.Brown has SIFT descriptors without affine information, so we use spatial NM-Net (nm-net-sp) instead.

As shown in Tab. \ref{tab:overallPreformance}, our method behaves favorably to both geometry and learning based methods on F1-score, especially on COLMAP and St.Brown, which are the challenging datasets with extremely low initial inlier ratios (7.50\% and 7.59\% respectively). 
For MSE measurement, all deep learning based methods obtain E matrix with similar precision.

Fig .\ref{visualize} shows the visual matching results of GLA-Net, our baseline (LFGC-Net) and classic RANSAC algorithm.

\section{Conclusion}
We present an end-to-end network for removing the wrong matches from the putative match set. In this network, we propose a novel Guided loss to establish the direct connection between loss and measurement, and an IA Block as feature extraction backbone to eliminate the impact of outlier on global context. We conduct extensive experiments to analyze the performance of each module and the overall network framework in detail. These experiments demonstrate that GLA-Net behaves favorably to the state-of-the-art approaches.

\bibliography{reference}
\bibliographystyle{aaai}
\end{document}